\documentclass{article}

\usepackage{spconf,amsmath,graphicx}

\usepackage[utf8]{inputenc} 
\usepackage[T1]{fontenc}    
\usepackage{hyperref}       
\usepackage{url}            
\usepackage{booktabs}       
\usepackage{amsfonts}       
\usepackage{nicefrac}       
\usepackage{microtype}      
\usepackage{xcolor}

\title{DSRN: An Efficient Deep Network for Image Relighting}

\name{Sourya Dipta Das$^{1,*}$ \qquad Nisarg A. Shah$^{2,*}$ \qquad Saikat Dutta$^{3}$ \qquad Himanshu Kumar$^{2}$}

\address{$^{1}$ Jadavpur University, India \quad
    $^{2}$ IIT Jodhpur, India \quad $^{3}$ IIT Madras, India \\
    \textit{$^*$Equal contribution}}

\usepackage{graphicx}
\begin{document}

\maketitle

\begin{abstract}
Custom and natural lighting conditions can be emulated in images of the scene during post-editing. Extraordinary capabilities of the deep learning framework can be utilized for such purpose. Deep image relighting allows automatic photo enhancement by illumination-specific retouching. Most of the state-of-the-art methods for relighting are run-time intensive and memory inefficient. In this paper, we propose an efficient, real-time framework Deep Stacked Relighting Network (DSRN) for image relighting by utilizing the aggregated features from input image at different scales. Our model is very lightweight with total size of about $42$ MB and has an average inference time of about $0.0116$s for image of resolution $1024 \times 1024$ which is faster as compared to other multi-scale models.  Proposed method is quite robust for translating image color-temperature from input image to target image. The method also performs moderately for light gradient generation with respect to the target image. Additionally, we demonstrate that the results further improve when images illuminated from opposite directions are utilized as input.

%


\end{abstract}

\section{Introduction}
Appropriate illumination is essential to obtain the desired images of a scene. But this cannot be achieved efficiently as one has very little or no control over the lighting. However, images can be modified to emulate the desired illumination conditions of the scene after the imaging process. There exist many digital tools for illumination manipulation but most of the tools have limited ability to manipulate the for small fluctuations in intensity or color. Whereas, the state-of-the-art \cite{wang2020deep, carlson2019shadow, gafton20202d} methods for illumination editing which are based on Deep learning framework, have much better illumination-editing capabilities compared to professionally available tools. However these  methods are computationally and memory intensive. This paper aims to develop an effective, lightweight framework for illumination editing that can also be deployed in the low-resource computing devices such as mobile. 

The image of a scene for a given illumination depends upon the various factors such as number and position of sources, surface properties, image plane position, and properties of illumination.  The radiant energy $I_p$ due to the point source $I_s$ at a given point at distance $r$ is given as in Equation \ref{eq:pointsource}.

\begin{equation}
    I_{p}  = \frac{I_s}{4\pi r^2}\hat{r}
\label{eq:pointsource}\end{equation}

Similarly, interaction of light with objects depends upon the properties of the surface such as orientation, texture and colour. The Phong model \cite{phong1975illumination} given in Equation \ref{eq:phong} can be utilized to describe the reflection of the various surfaces in such an illumination conditions. Thus, the image can appropriately be re-targetted for the cases when the illumination source is shifted to other locations. 

\begin{equation}
    I  = I_{p}K_{d}\hat{N}.\hat{L} + K_{a}I_{a} 
\label{eq:phong}\end{equation}

Here, $K_{d}$ is surface diffuse reflectivity, $I_{p}$ is point source intensity,  $\hat{N}$ is surface normal,  $\hat{L}$ is the light direction,  $K_{a}$ is the ambient reflectivity and $I_{a}$ is the ambient light intensity. In this work, we have considered only the diffused part of reflected light. Shadows are also one of the most important generated features  due to the light source position. Previous methods such as \cite{philip2019multi} produce the most aesthetic result, but are dependent on 3D models and priors for ascertaining the shadows and casting of shadows from the direction of the new light source. 

Here, the problem description is to transfer the illumination of given image from one fixed set of settings i.e. illumination direction and color temperature of the light source, to another fixed set of illumination settings of the target image of the same scene. In this paper, we refer to this case as single view problem. Along with the single view problem, we also demonstrate a study on  multi-illumination approach where images having same colour temperature but opposite illumination directions are utilized. These images are fused together to create a new average input image. This inturns is utilized as input in proposed method to reconstruct the output having the target illumination settings. We propose a fast and efficient network i.e.  Deep Stacked Relighting Network (DSRN) for both of the cases. We have done two separate studies on two approaches and have concluded that using two stage training process improves the performance of the model. Also, using multi view approach to this relighting problem additionally solves lot of discrepancies present in the results of single view approach.

\section{Related Work}\label{rwork}
Deep learning based methods have played a vital role in relighting real-life and synthetic scenarios with great efficiency. Gafton \emph{et al.} \cite{gafton20202d} proposed a GAN based Image translation methods like pix2pix \cite{isola2017image}. There is no additional knowledge requirement of the view geometry and parameters to generate proper lighting effects in this type of approach. Zhou \emph{et al.} \cite{zhou2019deep} proposed the use of Spherical Harmonics lighting parameters \cite{sengupta2018sfsnet} in the bottleneck layer of the UNet hourglass network to achieve state-of-the-art results over the portrait relighting task.

Ren \emph{et al.} \cite{ren2015image} introduces a regression-based neural network for relighting real-world scenes from a small number of images. We infer from these methods that the relighting problem can be reformulated as a image color translation from input image to target image and relighting based upon the explicit geometry information is computation intensive. Xu \emph{et al.} \cite{xu2018deep} proposed a deep learning-based framework to generate images under novel illumination using only five images captured under predefined directional lights. Their framework included a fully convolution neural network to predict the relighting function and input light direction.

Murmann \emph{et al.} \cite{murmann2019dataset} introduced a new multi-illumination dataset of real scenes captured under varying lighting conditions in high dynamic range and high resolution. They also proposed the use of UNet \cite{ronneberger2015u} for single-image relighting, treating as an image-to-image translation problem. It is a general observation that a minor change in lighting conditions alters many features in a scene and leads to performance degradation of this deep neural network based approach.

The approach proposed by Carlson \emph{et al.} \cite{carlson2019shadow} shows an application to relight outdoor scenes by adding a realistic shadow, shading, and other lighting effects onto an image used for tasks such as detection, recognition, and segmentation in urban driving scenes. This approach would be beneficial to make models for such tasks more robust. The notion of geometry can be encoded much more  efficiently in the encoder-decoder framework and by usage of multiple views of the scene. We encapsulate this information by using the stacked deep architecture as opposed to state of the art methods.   
\section{Proposed Approach}\label{proposed}
\textbf{Base Network:} We use a multi-scale network which works on a three-level image pyramid. In each level of the network, we have an encoder and a decoder. Decoder output of each level is upscaled and added to input image in next level and passed to encoder of next level. To capture global contextual information better, encoder output of each level is added with encoder output of previous level before feeding it to decoder. 

Let the input of encoder at $i$-th level $Encoder_i$ be $in_i$ and output of decoder at $i$-th level $Decoder_i$ be $out_i$. Hence, $in_i$ and $out_i$ are given by,

\begin{align}
    in_i = I_i + up(out_{i+1})\\
    F_i = Encoder_i(in_i)\\
    G_i = F_i + up(G_{i+1})\\
    out_i =  Decoder_i(G_i)
\end{align}
where, $I_i$ is the image at $i$-th level of image pyramid.  For the bottom-most level ($l$), terms $G_{l+1}$ and $out_{l+1}$ are absent. 

\begin{figure}[!h]
    \centering
    \includegraphics[scale=0.26]{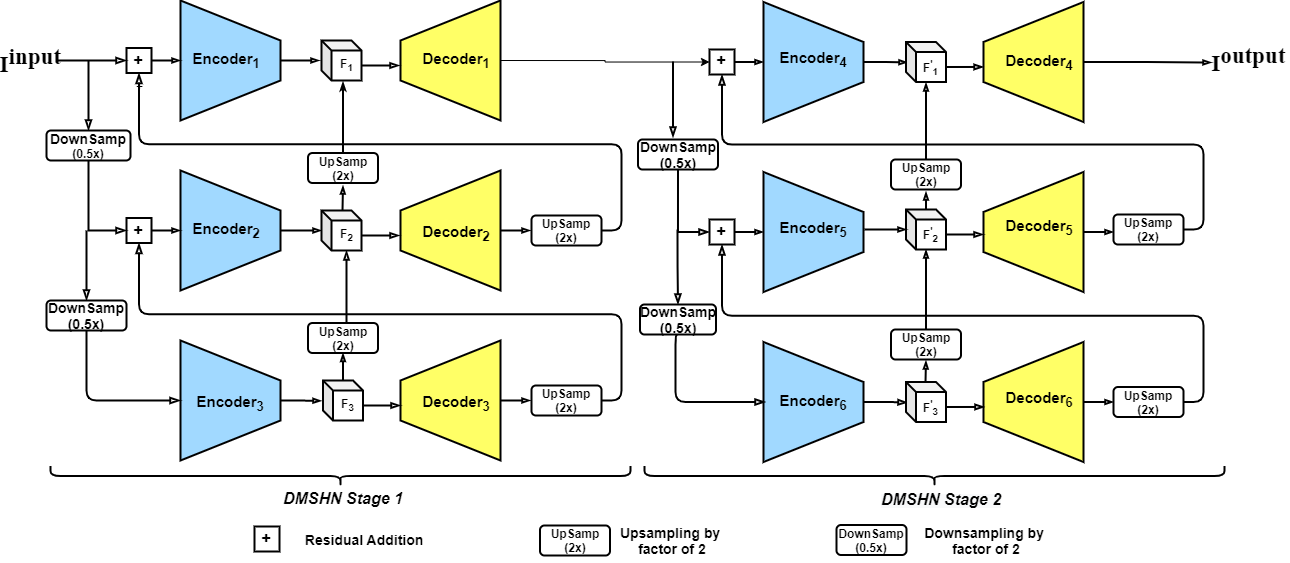}    
    \caption{Architecture Diagram of proposed DSRN} 
    \label{fig:my_label}
\end{figure}

We cannot improve performance of our model by adding additional lower levels in the base network as discussed in \cite{Zhang_2019_CVPR}. Thus, we cascade the same network twice to increase the performance of our network. Output of the first base network is fed to the second base network to generate the final output. We refer our final model as Deep Stacked Relighting Network (DSRN). The model architecture diagram is shown in Fig. \ref{fig:my_label}. 

\begin{figure}
    \centering
    \includegraphics[scale=0.27, angle=90]{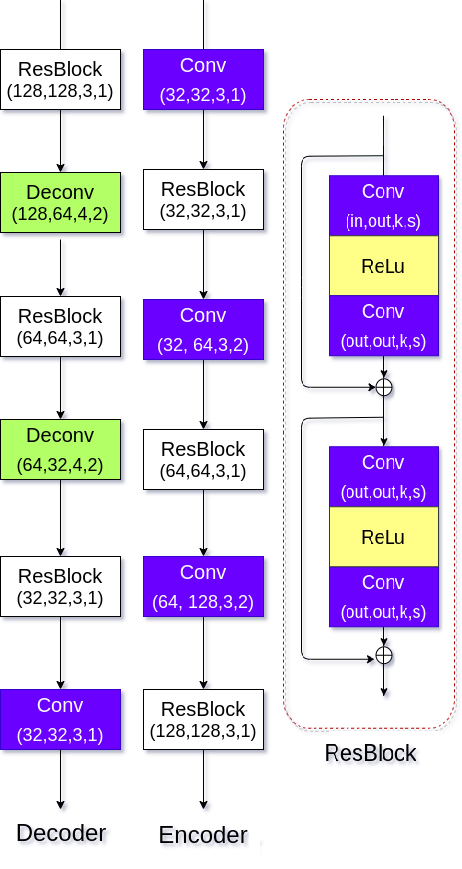}
    \caption{Architecture diagram of Encoder and Decoder. }
    \label{fig:enc-dec}
\end{figure}

Fig. \ref{fig:enc-dec} shows details of encoder and decoder architecture. Both encoder and decoder have a three-level hierarchy. Each hierarchy contains two residual blocks. Strided convolution is used for reducing spatial dimension in encoder and Transposed Convolution is used for increasing spatial dimension in decoder.

\textbf{Single-illumination Approach:}
\label{sec:single}
In case of Single illumination case, images in which illumination in the scene is coming from one particular direction and have a particular color temperature are taken as input. For target images, images  corresponding to different colour temperature and direction of illumination sources are selected.


\textbf{Multi-illumination Approach:}
\label{sec:multi} It is difficult to recover details in the dark regions of a single input image. On the other hand, if we have another input image of the same scene illuminated from opposite direction, most regions in the scene will appear to be bright in at least one of the input images. In multi-illumination approach, we leverage this extra information for image relighting.
In case of multi-illumination, two images where illumination is coming from two opposite direction in the scene having same colour temperature, are weighted averaged together to get input overlayed image for the model. Target image for this case is same as for the single-illumination case.
\begin{equation}
    I^{input}_{Avg} = W_1*I^{input}_{(D,T)} + W_2*I^{input}_{(D^*,T)}
\end{equation}
Here, $I^{input}_{(D,T)}$  is an image having illumination direction $D$ and color temperature $T$, $I^{input}_{(D^*,T)}$ is an image having opposite illumination direction $D^*$ and color temperature $T$, and $I^{input}_{Avg}$ is input overlayed or average image. Initially for our experiments, we have used $T = 6500$ for input image, and $D$  and $D^*$ are North (N) and South (S) directions respectively. The weight values $0.5$ is used for both  $W_1$ and $W_2$. 
\begin{figure}
    \centering
    \includegraphics[width=8.50cm,height=0.90in]{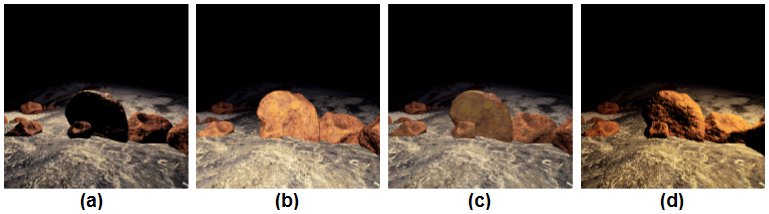}
    \caption{Examples of images at different stages during preprocessing for multi-illumination approach. From left: (a) Input image, $I^{input}_{(D,T)}$ , (b) Input image, $I^{input}_{(D^*,T)}$, (c) Processed weighted average or overlayed image, $I^{input}_{Avg}$, (d) Target Image}
    \label{fig:example}
\end{figure}

\textbf{Loss function:} We have done two stage training to enhance our results. First, we are using only $L_2$ loss as a reconstruction loss to train our model. Then, we used a linearly weighted loss function as objective loss function in second phase of training to get our final trained model. The combined loss function, $L_{CL}$ is given by,
\begin{equation}
    L_{CL} = \lambda_1*L_{1} + \lambda_2*L_{SSIM} + \lambda_3*L_p + \lambda_4*L_{tv}
\end{equation}
where, $L_{1}$ is Mean Absolute Error (MAE) loss, $L_{SSIM}$ is SSIM loss \cite{wang2004image} , $L_p$ is  Perceptual loss \cite{gatys2016image} and $L_{tv}$ is TV loss \cite{mahendran2015understanding}. During training, values of $\lambda_1$, $\lambda_2$, $\lambda_3$ and $\lambda_4$ are chosen to be $1$, $5 \times10^{-3}$, $6 \times 10^{-3}$ and $2\times10^{-8}$ respectively.
\vspace{-10px}
\begin{table}[!h]
\caption{Performance of the proposed relighting frameworks for various loss functions.}
\small
\centering
\tabcolsep=0.11cm
\begin{tabular}{|c|c|c|c|}
\hline
\textbf{Method} & \textbf{PSNR $\uparrow$} & \textbf{SSIM $\uparrow$} & \textbf{LPIPS $\downarrow$} \\ \hline
Base Network ($L_{MSE}$) & 16.94 & 0.5659 & 0.4933 \\ \hline
Base Network ($L_{CL}$)  & 17.20 & 0.5696 & 0.3712 \\ \hline 
DSRN ($L_{MSE}$) & 17.53 & 0.5673 & 0.4253 \\ \hline
DSRN ($L_{CL}$) & 17.89 & 0.5899 & 0.4088 \\
\hline

\end{tabular}
\label{tab:singleview}
\end{table}

We optimized the proposed DSRN by employing the performance test with different loss functions and architectures.
We utilized $L_2$ loss $(L_{MSE} )$ and Combined Loss $(L_{CL})$ to train our networks.
Base and Stacked Networks trained with different loss functions as shown in Table \ref{tab:singleview}. We observe from the table that the proposed stacked model DSRN provides optimal performance for Combined Loss $(L_{CL})$ function.  



\section{Results and Discussion}\label{results}

\quad\textbf{System Description:} Our performance evaluation framework includes Pytorch in Linux environment with AMD $1950$X processor along with $64$GB RAM and $11$ GB NVIDIA GTX $1080$ Ti GPU.  In all of our experiments, an initial learning rate of $2\times10^{-3}$ is selected which is gradually decreased to $5\times10^{-5}$. During training, we kept the batch size as $2$ and  input image size as $512\times 512$.

\textbf{Dataset Details:} We have utilized VIDIT dataset \cite{helou2020vidit} for our experiments. The dataset contains $15600$ images from $390$ different scenes in $40$ different illumination settings. The scenes are illuminated with $5$ sources of different color temperatures from $8$ azimuthal angle directions one at a time. We have used $300$ scene images of size $512 \times 512$ for training and $45$ scene images for validation and testing each. For  multi-illumination approach, we need  multi-illumination data for both training and validation but as  multi-illumination data for validation set is not available, we split up our training data into two sets of  custom training and custom test set. For custom test set\label{data:test}, we sampled 60 images randomly from whole training data.

\textbf{Evaluation:} We report the performance comparison over validation dataset as ground truth for test set is not made publicly available. We compare the performance using metrics PSNR, SSIM \cite{ssim} and LPIPS \cite{zhang2018unreasonable}.
\vspace{-10px}
\begin{table}[!h]
\caption{Performance comparison of proposed DSRN with state of the art methods on \cite{helou2020vidit}.}
\small
\centering
\tabcolsep=0.11cm
\begin{tabular}{|c|c|c|c|c|}
\hline
\textbf{Method} & \textbf{PSNR} $\uparrow$ & \textbf{SSIM} $\uparrow$ & \textbf{LPIPS} $\downarrow$ & \textbf{Runtime (s)} \\ \hline
Pix2Pix \cite{isola2017image} & 16.28 & 0.553 & 0.482 & 0.2504 \\ \hline
Dense-GridNet \cite{liuICCV2019GridDehazeNet} & 16.67 & 0.2811 & 0.3691 & 0.9326 \\ \hline
SRN \cite{tao2018scale} & 16.94 & 0.5660 & 0.4319 & 0.8710 \\ \hline
DRN \cite{wang2020deep} & 17.59 & 0.596 & 0.440 & 0.5012 \\ \hline
DSRN (Proposed) & 17.89 & 0.5899 & 0.4088 & 0.0116 \\ \hline
\end{tabular}
\label{tab_other_methods}
\end{table}

Table \ref{tab_other_methods} presents the performance comparison of the proposed DSRN framework for relighting with the state of the art methods \cite{liuICCV2019GridDehazeNet}, \cite{tao2018scale} and \cite{wang2020deep}. We observe from the table that the proposed DSRN based relighting framework produces more accurate output with more than $25$x speedup in the computation time. This highlights the efficacy of the proposed DSRN for the real-time relighting.  

\begin{figure*}[h]
    \centering
    \includegraphics[width=0.8\textwidth]{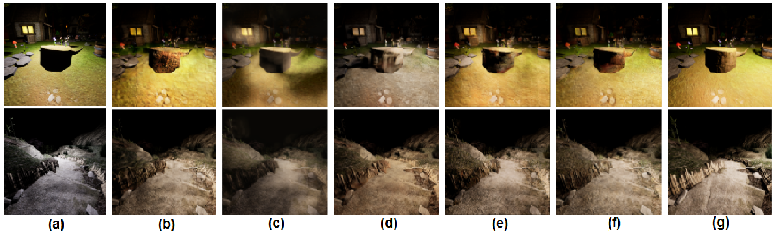}
    \vspace{-10px}
    \caption{Qualitative comparison of proposed DSRN with state of the art methods. From left: (a) Input Image ($I^{input}_{(D,T)}$), (b) SRN \cite{tao2018scale}, (c) Dense-GridNet \cite{liuICCV2019GridDehazeNet}, (d) DRN\cite{wang2020deep}, (e) Base Network (Proposed), (f) DSRN (Proposed), (g) Ground Truth.}
    \label{fig:other_model}
\end{figure*}

\begin{figure*}[!h]
    \centering
    \includegraphics[width=\textwidth, height=0.8in]{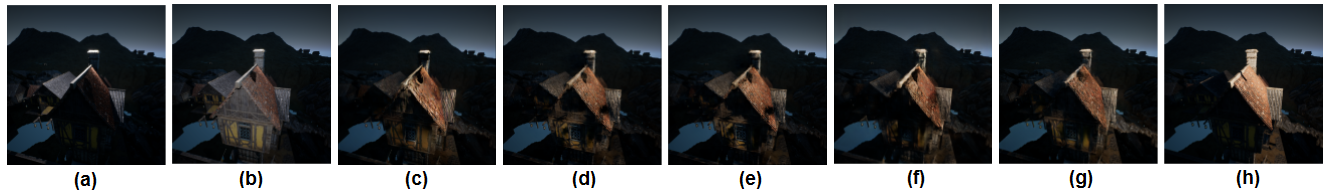}
    \vspace{-20px}
    \caption{Qualitative Comparison of our method for Multi-illumination Approach against Ground Truth on custom test data. From left: (a) Input Image ($I^{input}_{(D,T)}$), (b) Input Image ($I^{input}_{Avg}$), (d) Base Network with $L_2$ Loss, (e) Base Network with Combined Loss, (f) DSRN with $L_2$ Loss, (g) DSRN with Combined Loss, (h) Ground Truth.} 
    \label{fig:multiview_results}
\end{figure*}

We also present the qualitative comparison of image relighting of proposed DSRN framework with state-of-the art methods as shown in the Fig. \ref{fig:other_model}. We observe from the figure that the proposed method produces relighting output of higher quality compared to state of the art methods \cite{liuICCV2019GridDehazeNet}, \cite{tao2018scale} and \cite{wang2020deep} which suffer from the various relighting-artifacts. 



We have also extended the proposed DSRN framework for multi-illumination case to improve the results in the saturation regions. We evaluated the performance of the multi-illumination based approach on custom dataset. Table \ref{tab:multiview}  shows the performance metrics for the multi-illumination based image relighting frameworks. We observe from the Table \ref{tab:singleview} and  \ref{tab:multiview} that DSRN leverages the most from the additional information and results in significant improvement in PSNR, SSIM and LPIPS. This validates the efficacy of the proposed DSRN for image relighting task.
\vspace{-10px}
\begin{table}[!h]
\centering
\caption{Performance metrics for the multi-illumination based relighting using proposed DSRN  on \cite{helou2020vidit}.}
\label{tab:multiview}
\small
\begin{tabular}{|c|c|c|c|}
\hline
\textbf{Method } & \textbf{PSNR $\uparrow$} & \textbf{SSIM $\uparrow$} & \textbf{LPIPS $\downarrow$} \\ \hline
Base Network ($L_{MSE}$) & 17.88 & 0.6372 & 0.2547 \\ \hline
Base Network ($L_{CL}$)  & 17.87 & 0.6445 & 0.2532 \\ \hline
DSRN ($L_{MSE}$) & 19.02 & 0.6725 & 0.3913 \\ \hline
DSRN ($L_{CL}$)  & 19.25 & 0.7038 & 0.3265 \\ \hline
\end{tabular}
\end{table}
We also present the qualitative performance improvement in the image relighting results due to usage of the multi-illumination data as shown in the Fig. \ref{fig:multiview_results}. We observe from the figure that multi-illumination data improves the quality of relighted images in the saturation regions.

We also tested the performance of the multi-illumination based image relighting for various combinations of the illumination and target illumination as given in Table \ref{tab:multiview_extra}. We observe from the table that the proposed DSRN is robust with respect to different illumination conditions. Also, illumination combination covering the entire depth of the scene (N and S) is usually more preferable compared to other combination.  
Thus, we conclude from the results that the proposed framework of DSRN is an effective framework for real-time image relighting. The utilization of multi-illumination data further improves the performance of the proposed framework.  
\vspace{-10px}
\begin{table}[!h]
\centering
\caption{Performance metrics for proposed relighting using DSRN for multi-illumination conditions.}
\label{tab:multiview_extra}
\small
\begin{tabular}{|c|c|c|c|c|}
\hline
\begin{tabular}[c]{@{}c@{}}\textbf{Source }\\\textbf{Settings } \end{tabular} & \begin{tabular}[c]{@{}c@{}}\textbf{Target }\\\textbf{Settings } \end{tabular} & \textbf{PSNR $\uparrow$} & \textbf{SSIM $\uparrow$} & \textbf{LPIPS $\downarrow$} \\ \hline
N and S & E  & 19.25 & 0.7038 & 0.3265 \\ \hline
N and S & W  & 19.55 &  0.6955 & 0.2860 \\ \hline
E and W & S  & 19.55 & 0.6797 & 0.2839 \\ \hline
E and W & N  & 17.53 &  0.6556 & 0.3169 \\ \hline
\end{tabular}
\end{table}
\vspace{-15px}
\section{Conclusion}\label{conclusion}
In this paper, we have proposed a fast and lightweight stacked network, DSRN for image relighting. The proposed network achieves state-of-the-art results with more than $25$x speedup. We have adopted a two-stage training process to improve results both quantitatively and qualitatively. Through extensive experiments, we have demonstrated the capability of the proposed DSRN in both single-illumination and multi-illumination settings. In future, we would like to explore these aspect to develop a more complete image relighting framework. Also, we would like to include the both diffused and specular components of the reflected illumination for better and more realistic scene relighting in the future.

\bibliographystyle{IEEEbib}
\bibliography{egbib}






\end{document}